\documentclass[times,twocolumn,final]{elsarticle}

\usepackage{template/cag}
\usepackage{framed,multirow}

\usepackage{amssymb}
\usepackage{latexsym}

\usepackage[hyphens]{url}
\usepackage{xcolor}
\definecolor{newcolor}{rgb}{.8,.349,.1}

\usepackage{hyperref}

\usepackage[switch,pagewise]{lineno} 

\usepackage{lipsum}
\usepackage{booktabs} 
\usepackage{amsmath}
\usepackage{amssymb}
\usepackage{enumitem}
\usepackage{textcomp}

\usepackage{makecell}
\usepackage{colortbl}
\usepackage{multirow}
\usepackage{tabularray}
\usepackage{color,soul}
\usepackage{ulem}
\usepackage{xspace}

\def\revisionmode{0} 

\newcommand{\revision}[2][]{%
\if\revisionmode1%
\if\relax\detokenize{#1}\relax%
\textcolor{blue}{#2}%
\else%
\textcolor{blue}{\hl{(\textbf{Note:} #1)} #2}%
\fi%
\else%
#2%
\fi%
}%

\newcommand{\removal}[1]{%
\if\revisionmode1%
\textcolor{red}{\sout{#1}}%
\fi%
}%

\newcommand{\revisionnote}[1]{%
\if\revisionmode1%
\hl{\textbf{Note:} #1}%
\fi%
}%

\journal{Computers \& Graphics}

\begin{document}

\verso{Author Accepted Version}

\begin{frontmatter}


\title{Towards mitigating uncann(eye)ness in face swaps via gaze-centric loss terms}
\author[1]{Ethan \snm{Wilson}}
\emailauthor{ethanwilson@ufl.edu}{Ethan Wilson}

\author[2]{Frederick \snm{Shic}}
\emailauthor{fshic@uw.edu}{Frederick Shic}

\author[3]{Sophie \snm{J\"{o}rg}}
\emailauthor{sophie.joerg@uni-bamberg.de}{Sophie J\"{o}rg}

\author[1]{Eakta \snm{Jain}}
\emailauthor{ejain@ufl.edu}{Eakta Jain}

\address[1]{Computer \& Information Science \& Engineering Department, University of Florida, USA}
\address[2]{University of Washington, USA}
\address[3]{University of Bamberg, Germany}

\received{\today}
\begin{abstract}

Advances in face swapping have enabled the automatic generation of highly realistic faces.  
Yet face swaps are perceived differently than when looking at real faces, with key differences in viewer behavior surrounding the eyes.
Face swapping algorithms generally place no emphasis on the eyes, relying on pixel or feature matching losses that consider the entire face to guide the training process.
We further investigate viewer perception of face swaps, focusing our analysis on the presence of an uncanny valley effect.  We additionally propose a novel loss equation for the training of face swapping models, leveraging a pretrained gaze estimation network to directly improve representation of the eyes.
We confirm that viewed face swaps do elicit uncanny responses from viewers.  Our proposed improvements significant reduce viewing angle errors between face swaps and their source material.  Our method additionally reduces the prevalence of the eyes as a deciding factor when viewers perform deepfake detection tasks.
Our findings have implications on face swapping for special effects, as digital avatars, as privacy mechanisms, and more; negative responses from users could limit effectiveness in said applications.  Our gaze improvements are a first step towards alleviating negative viewer perceptions via a targeted approach.

\end{abstract}

\received{6 September 2023}
\accepted{1 February 2024}

\end{frontmatter}


\if\revisionmode1
\section*{Revision Legend:}

I have created a few macros to revise the text and make changes clear.  You can see each type of revision below:

\begin{itemize}[noitemsep,topsep=0pt]
    \item \revision{This is a simple revision which is highlighted here and present in the text.}
    \item \revision[This is a note to specify the review comment being addressed and/or justify the change.  This does not show in the final text.]{This is a revision which is highlighted here and present in the text.}
    \item \removal{This indicates that some text has been removed, and is not present in the final text rendering.}
    \item \revisionnote{This is a standalone note to give information to reviewers, this is not present in the final text rendering.}
\end{itemize}
\fi

\section{Introduction}

Face swapping is the act of placing a \textit{character's} face overtop of an \textit{original} face in digital media.  Recent deep learning models capable of face swapping are able to generate highly realistic faces that are indistinguishable to naive viewers.  

Recent perceptual work has found that viewers perceive face swaps differently than real faces, with a large portion of differences revolving around perception of the eyes.  We hypothesize that this is due to how face swap models are trained;  the main optimization is to disentangle the identity from non-identity attributes, yet non-identity attribute matching is typically covered by broad image reconstruction loss.  While human viewers focus heavily on the eyes, these loss equations do not, and consider the full face equally.  We believe that because of this training paradigm, the generated eyes are lackluster.  We also hypothesize that they eyes could influence the presence of a potential uncanny valley effect when viewing face swaps.

We begin our analysis with an investigation into the uncanniness of face swaps.  Through a perceptual study, we show that face swaps generally elicit an uncanny feeling when viewed.  We then evaluate methods to improve the quality of generated eyes in face swaps.  Using predicted gaze angles of original and reconstructed faces, we define a reconstruction loss term focused on the eyes to add a gaze component to the overall loss function, enhancing the accuracy of reconstructed gaze without compromising visual quality.  We evaluate our method via a perceptual study consisting of deepfake detection and an uncanniness questionnaire.

Our explicit focus on preserving gaze behavior could be applied to existing or future face swapping pipelines.  Our implementation alters the optimization function but does not alter model architecture, meaning that already-trained models can be fine-tuned with this improvement in place.  We showcase the method on a popular open-source face swapping network, seeing significant improvement in reconstructed gaze directions compared to baseline face swapping and altering the prevalence of the eyes as a selective attribute in deepfake detection tasks.

\subsection{Extension of Previous Work}

\revisionnote{Moved this subsection up in order to make it clearer to readers initially that this is an extension of previous work.}

This work is a direct extension of the publication \textit{Introducing Explicit Gaze Constraints to Face Swapping}~\cite{wilson_introducing_2023}, presented at the \textit{ACM Symposium on Eye Tracking Research and Applications (ETRA) 2023}.  Preliminary work leading to the findings in Section~\ref{sec:uncanniness-of-face-swaps} was presented in poster format as \textit{The Uncanniness of Face Swaps}~\cite{wilson_uncanniness_2022} at the"\textit{Vision Sciences Society (VSS) Annual Meeting 2022} and subsequently published in the \textit{Journal of Vision} as an extended abstract.  \revision{At \textit{ETRA 2023}, we presented a novel loss equation to guide the training of face swap models while placing an explicit focus on the eyes, and showed significant improvements in gaze reconstruction error over a baseline architecture.  In this extended work,} we formally present evidence that face swaps are within the uncanny valley region, then extend our prior analysis of the proposed gaze improvements to qualitatively evaluate deepfake detection, attribute importance, and uncanniness.

\subsection{Ethics of Face Swapping}

Face swapping has uses in visual effects, interactions with virtual avatars~\cite{caporusso_deepfakes_2021}, and privacy protection~\cite{zhu_deepfakes_2020, lee_american_2021}; however, face swapping has become a controversial technology due to its potential for impersonation, spreading misinformation and violating individuals' privacy.  These so called \textit{deepfakes}' sudden accessibility has incited public concern and sparked legislative response~\cite{wagner_word_2019}.  Yet, responsible innovation on face swapping is necessary and will lead to positive outcomes.  The content this paper explores could increase the naturalness of future face swapping algorithms, making them more feasible in positive applications for social good, but also in negative contexts.  \revision[To address reviewer 1's concerns regarding our ethical stance, we have added some discussion on the current commercialization of this technology.  Essentially, there are high commercial incentives for face editing technologies.  Even if work in this space is not published, I believe it will be researched behind closed doors, potentially being patented or guarded as company secrets.  By committing to open science, these improvements to generated faces can also be taken advantage of in research to \textit{detect} generated faces.]{The number of emerging commercial applications of this technology\footnote{Examples include the use of face swapping and de-aging in blockbuster films such as Star Wars, the Fast franchise, and the Irishman, and synthesizing David Beckham's speech in 9 languages during the Malaria no More campaign (\url{https://www.synthesia.io/post/david-beckham}).} indicate that face-editing technologies will remain and continue to advance~\cite{meskys_regulating_2019}.  We believe that by openly publishing innovations in this field, we will be able to benefit the positive use cases of face swapping while preventing improper and malicious use.  The dreary alternative would be the commercialization of the underlying technologies, making publicly-available deepfake detection more difficult and less accessible.}  

These innovations will \removal{also} aid in future detection of deepfakes, \revision{helping to regulate the technology.}  Classifiers based on biometric signals, including gaze patterns, are being developed for deepfake detection~\cite{jung_deepvision_2020, ciftci_how_2020, demir_where_2021}.  These methods train on real and swapped face videos.  By feeding these models new training data with more believable gaze \revision{patterns,} we will narrow the decision boundary between real and fake stimuli, enabling further training and innovation, ultimately seeing increased accuracy and reliability when detecting fake media across the internet.

\subsection{Main Contributions}
 
This work identifies that face swaps generally do elicit uncanny feelings in viewers.  Given the prevalence of the eyes as noticed artifacts in prior perceptual studies regarding face swapping, we hypothesize that a large portion of the uncanniness felt by users could be attributed to the eyes.  We then evaluate methods which place explicit constraints on the eyes during training of face swapping models, better preserving gaze directions of the driving video.  Our proposed method is modular and could easily extend existing face swapping architectures.  We finally validate our method's effect on viewer perception, finding that our proposed improvements alter the reliability of gaze as a factor when determining if a face is real or fake.

\subsection{Roadmap}

In Section~\ref{sec:related-work}, we will review related literature, including the algorithmic improvements that have enabled face swapping to achieve a high level of realism and user perception of face swaps and other computer generated faces.  In Section~\ref{sec:uncanniness-of-face-swaps}, we will detail an experiment that verifies that users do experience uncanny feelings when viewing face swaps.  In Section~\ref{sec:methodology}, we will detail a novel loss formulation which enables face swaps to directly enforce the accuracy of reconstructed gaze during training.  We will also introduce a perceptual study which aims to evaluate the efficacy of the improved training paradigm against a baseline face swap and an alternative method to improve gaze.  In Section~\ref{sec:results}, we will evaluate our method both quantitatively via the accuracy of reconstructed gaze and qualitatively based on responses of the perceptual study.  Section~\ref{sec:discussion} will discuss our findings, their implications to broader applications, and limitations of the given evaluation.  Finally, we will conclude with Section~\ref{sec:conclusion}.
\section{Related Work}
\label{sec:related-work}

We begin this literature overview by presenting a summary of the algorithmic innovations that have enabled high-quality synthesized faces which are now very difficult for naive viewers to detect.  We then cover relevant perceptual insights that prior research has revealed regarding face swaps.  It has been hypothesized that face swaps could fall within the uncanny valley~\cite{lyu_deepfake_2020}, but there is no existing literature to support or reject this; we instead highlight a rich body of work assessing the uncanniness of computer generated avatars, as a key emerging use for face swapping is to generate custom virtual avatars.  We then discuss key insights regarding the importance of gaze, and why gaze may not be properly represented in existing face swapping approaches.

\subsection{Algorithmic Innovations}

Recent innovations in image generation techniques, most prominently the generative adversarial network (GAN)~\cite{goodfellow_generative_2014}, variational autoencoder (VAE)~\cite{kingma_auto-encoding_2022}, diffusion probabilistic model~\cite{ho_denoising_2020} and key advancements thereafter~\cite{karras_analyzing_2020, razavi_generating_2019, radford_unsupervised_2016, liu_coupled_2016, rombach_high-resolution_2022}, have rapidly advanced the ability to create realistic AI-synthesized faces.  These technologies paved the way for powerful, fully automated face swaps that have become nearly undetectable to naive human viewers.  

Face swapping has become an active area of research in recent years~\cite{zhang_deepfake_2022, dang_digital_2023, walczyna_quick_2023}.  The first deep-learning specialized face swapping algorithm is a forked autoencoder with two distinct decoders, each training on a unique identity\footnote{This algorithm was originally shared by Reddit user /u/deepfakes in 2017; the project is still actively maintained as "faceswap": \url{https://github.com/deepfakes/faceswap}}~\cite{noauthor_faceswap_nodate}.  Advancements over this initial method have focused on swapping between arbitrary identities~\cite{nirkin_fsgan_2019, li_advancing_2020, chen_simswap_2020, liu_blendgan_2021}, real-time applications~\cite{korshunova_fast_2017, thies_facevr_2018}, and achieving higher resolutions~\cite{liu_deepfacelab_2023, zhu_one_2021, wang_hififace_2021} and pose consistency~\cite{li_3d-aware_2023, nitzan_face_2020}.  There are multiple face swapping platforms online\footnote{https://github.com/shaoanlu/faceswap-GAN}\footnote{https://github.com/iperov/DeepFaceLab}\footnote{https://github.com/deepfakes/faceswap/}, making this technology accessible for VFX artists, hobbyists, and researchers.

Some image and face synthesis methods have begun to leverage existing networks, hereafter referred to as \textit{pretrained expert models}, as part of their training process.  Facial recognition systems~\cite{deng_arcface_2019, wang_cosface_2018} have been used to automatically segment identity or to obtain an overall attribute profile~\cite{chen_simswap_2020, tang_cycle_2019, nitzan_face_2020}; facial attribute extractors have been used to classify the face in an unsupervised manner~\cite{li_anonymousnet_2019, xue_face_2023}; landmark estimators have been used to extract or enforce body or facial structure~\cite{sun_natural_2018, nitzan_face_2020, kuang_effective_2021, siarohin_motion_2021}; style transfer algorithms extract style using pretrained networks'~\cite{simonyan_very_2015} intermediate features~\cite{liu_blendgan_2021, gatys_image_2016, zhang_multi-style_2018, huang_arbitrary_2017, wang_high-resolution_2018}.  These methods found success using high-level predictions or intermediate features from pretrained expert models to aid in training without requiring supervised labels.

\subsection{Perception of Face Swaps}

The perception of face swapped videos has been explored as a way to aid in detection efforts against fake media~\cite{preu_perception_2022}. Recent work has investigated the accuracy with which humans can guess if a video is genuine (original/real) or face swapped (manipulated/fake)~\cite{rossler_faceforensics_2019, tahir_seeing_2021, groh_deepfake_2022, groh_human_2023}, what artifacts they pick up on~\cite{wohler_pefs_2020}, if the emotions in those videos are rated as relatively insincere~\cite{wohler_towards_2021}, and the trustworthiness of generated faces~\cite{wohler_personality_2022, nightingale_ai-synthesized_2022}. 

In contrast to the emerging literature on the perception of face swaps, there is a rich body of work on the perception of computer-generated (CG) characters and faces~\cite{mcdonnell_render_2012, hodgins_saliency_2010, carter_unpleasantness_2013, carrigan_investigating_2020}.  As virtual characters became more and more realistic, research focusing on the uncanny valley effect arose \cite{macdorman_too_2009, geller_overcoming_2008, ho_measuring_2017}. The uncanny valley can explain the feeling of discomfort elicited when viewing a synthetic face that is similar to a real face but not \textit{quite right}.  Originally an observation of human-like robots~\cite{mori_uncanny_2012}, the uncanny valley theory has expanded from robots to other media, such as virtual characters in images and videos.  

CG faces in images have been shown to have a more negative eeriness effect when compared to human-created faces along the same human-likeness indices~\cite{katsyri_virtual_2019}.  A justification for the higher negative reaction to CG faces is a high likelihood in mismatched realism across different facial features~\cite{macdorman_categorization-based_2017}, which has been shown to increase negative affinity~\cite{macdorman_reducing_2016}.  Additionally, human-like CG characters have been found to have more of an uncanny effect in video segments compared to still images~\cite{dill_evaluation_2012}.  In video, the uncanny valley effect has also been shown to be present in inconsistent or degraded motions~\cite{white_motion_2007, piwek_empirical_2014}.

There is little work relating the uncanny valley effect to face swaps.  \revision{While W\"{o}hler et al.~\cite{wohler_personality_2022} measure eerieness and appeal as part of a broader study on personality, uncanniness is not directly addressed or evaluated in-depth.}  Yet, in the field of deepfake detection, biometric-focused detection methods have seen success~\cite{ciftci_fakecatcher_2020, ciftci_how_2020, demir_where_2021, li_ictu_2018}.  These methods rely on temporal features such as eye movements, blink patterns, or facial expressions, using irregularities to classify videos as fake.  It is clear that there are inconsistent motions between attributes of real and swapped faces, indicating that face swaps may fall within the uncanny valley~\cite{mullen_new_2022}.

\subsection{Importance of Gaze}

The core goal of modern face swapping is to disentangle the embedded feature vector between identity and other facial attributes, so that identities can be swapped while all other features remain constant.  While each algorithm is unique, in nearly all methods the problem is framed as \textbf{identity} \textit{versus} \textbf{all attributes}, i.e. all aspects outside of identity are placed under a single loss term.  For example, multiple approaches isolate and replace the identity portion of a feature embedding~\cite{korshunova_fast_2017, chen_simswap_2020, li_advancing_2020, wang_hififace_2021} or bake identity into the weights of the model~\cite{liu_deepfacelab_2023, noauthor_faceswap_nodate}. The overall attribute profile is preserved, but is enforced only according to a general image-based reconstruction loss, which may fail to emphasize perceptually relevant features.  Particularly, the eyes spatially occupy only about 5.6\% of the face, yet human viewers focus on the eyes approximately 40\% of the time~\cite{janik_eyes_1978}.  Because features are derived implicitly from pixel images, the eyes are not prioritized, thus have been found to account for a large percent of noticed artifacts~\cite{wohler_towards_2021, tahir_seeing_2021, gupta_eyes_2020}.  

A simple way to mitigate this shortcoming is through brute force --- create a deeper network with a larger latent space.  For example, using eight identity-specific decoders rather than two and increasing model depth~\cite{naruniec_high-resolution_2020}.  This is effective yet sees increased training times and memory requirements.  \revision{Instead, we place an explicit focus on the eyes through targeted loss equations, enforcing the model to prioritize the eyes in a way more aligned with human perception.}  \revisionnote{Reviewer 1's understanding was that our compared DFL+em condition is a method that has been proposed and evaluated.  Actually, em is included within Deepfacelab's codebase (at https://github.com/iperov/DeepFaceLab), with a parameter to enable or disable the term.  However, this method had no prior evaluation and is only mentioned in passing in their publication, stating that weighing different parts of the face is a "useful trick".  I am removing this mention from the related works and better explaining the status of our em comparison method when it is introduced in Section 4.1.} \removal{Instead of increasing resources, another potential solution is to add a gaze-aware constraint to the training process, but this method is not well explained and lacks any evaluation in the corresponding manuscript~\cite{liu_deepfacelab_2023}.}

\section{Evaluating the Uncanniness of Face Swaps}
\label{sec:uncanniness-of-face-swaps}

In prior dealings with face swapping, we had noticed that faces tended to look \textit{off-putting}~\cite{wilson_uncanniness_2022}.  Prior work has perceptually analyzed face swaps, finding key differences between viewing behaviors when analyzing real or generated faces~\cite{wohler_towards_2021, tahir_seeing_2021, gupta_eyes_2020}.  We hypothesize that this viewing phenomenon could be explained under the lens of the \textit{uncanny valley effect}~\cite{mori_uncanny_2012}.  The uncanny valley effect hypothesizes a relation between a character's level of human features and the emotional response elicited from human viewers.  In the space between \textit{partially} and \textit{fully} human, there is a valley where viewers become uncomfortable, repulsed, and/or unempathetic as the character fails to accurately mimic human features.


We design a perceptual study to evaluate the state of face swapping technology and its relation to \textit{uncanniness} and how these synthetic videos relate to genuine videos that do not contain face swaps.  Our hypothesis is:

\begin{itemize}
    \item H1: Face swapped videos are perceived as more \textbf{uncanny} than original videos.
\end{itemize}

\subsection{Stimuli} 

\begin{figure}[h]
    \centering
    \includegraphics[width=1\linewidth]{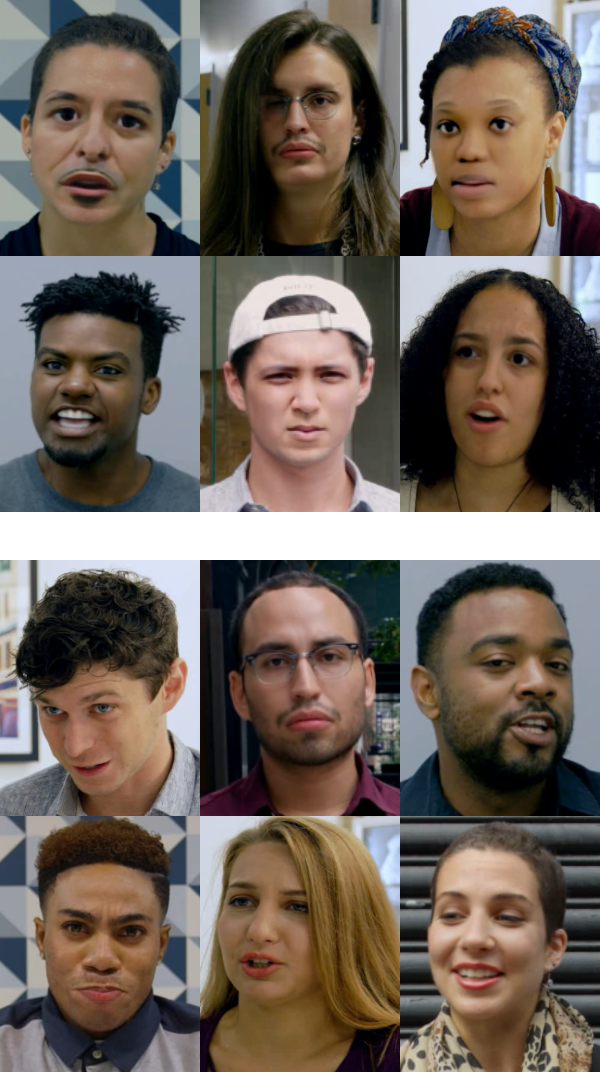}
    \caption{\revisionnote{Expanded number of faces.} A selection of stimuli from FF++ DFD used to evaluate uncanniness.  Top = face swaps; Bottom = real faces.}
    \label{fig:uncanniness-thumbnails}
\end{figure}

We extract 40 videos from the FaceForensics++ Deepfake Detection dataset (FF++ DFD)\footnote{https://ai.googleblog.com/2019/09/contributing-data-to-deepfake-detection.html}~\cite{rossler_faceforensics_2019}. There are 20 face swap videos and 20 original videos, chosen as the first videos returned when querying the dataset.  Each video is trimmed to the median 10 seconds to standardize video lengths, decrease study runtime, and eliminate entrance/exit transitions common in many clips.  A selection of example faces from the video clips can be seen in Figure~\ref{fig:uncanniness-thumbnails}

\subsection{Participants} 

Survey participants were recruited under IRB approved protocol via several communication channels including word of mouth and electronic mailing list advertisements ($N=39$; 59\% male, 38\% female, 2.5\% other).  The survey population consists mainly of undergraduate University students. The racial-ethnic distribution is 77\% White, 18\% Asian, 5\% Black or African American, 2.5\% Native American and 2.5\% other, where 5\% of participants report two or more races. 18\% report being Hispanic/Latinx. The median age is 21 years (IQR = 20-23.5). Survey data is anonymized for subsequent analysis.

\subsection{Procedure}
\label{sec:uncanniness_protocol}

The survey was conducted online and hosted via Qualtrics\footnote{https://www.qualtrics.com}. Prior to taking the survey, the participants were given a definition of face swapping and informed that they may encounter face swaps. However, they were not informed of whether each video was a face swap or not. Participants were then shown one stimulus video at a time and asked to watch each video in its entirety. Participants were then asked to rate the stimulus on 7-point Likert scales between five pairs of bipolar adjectives. 

The adjective pairs we have employed are a modified subset of the pairs used by Ho and MacDorman~\cite{ho_measuring_2017} in their semantic differential scales to measure three categories of uncanniness: \textit{humanness, eeriness,} and \textit{attractiveness}.  The pair \textbf{Real/Synthetic} relates to humanness, \textbf{Agreeable/Repulsive} relates to attractiveness, and \textbf{Plain/Weird} relates to eeriness.  We define \textbf{Ordinary/Uncanny} as a pairing using adjectives sampled from two pairs (Ordinary/Supernatural, Bland/Uncanny) from within the eeriness category.  \textbf{Unremarkable/Unusual} proposes new adjectives, but is a similar pairing to Uninspiring/Spine-tingling and Unemotional/Hair-raising, both in the eeriness category.  Our choice in adjective pairs cover the full spread of Ho and MacDorman's proposed categories with a large emphasis on eeriness.  We average the responses on the five adjective pairs to obtain an overall assessment of the uncomfortable feeling a viewer has, in other words, the uncanniness of the presented video.  A score of one on the Likert scale represents the words closest to \textit{normal}, while seven represents the words analogous to \textit{uncanny}.  Each participant viewed each stimulus; presentation order was randomized per participant to account for an anticipated order-effect bias \cite{schuman_questions_1996}.  

\subsection{Results}

\begin{table}[th!]
\centering
    \begin{tabular}{@{}l l l l l l l@{}}\toprule
        & \multicolumn{2}{c}{Original} && \multicolumn{2}{c}{Face Swap} & \\
        \cmidrule{2-3} \cmidrule{5-6}
        Adjective Pair & Mean & SD && Mean & SD \\ 
        \midrule
            Real/ \\ Synthetic          & 2.11 & 0.99 && 5.20 & 0.90 \\ \midrule
            Agreeable/ \\ Repulsive     & 2.18 & 0.95 && 4.49 & 0.96 \\ \midrule
            Unremarkable/ \\ Unusual    & 2.26 & 0.90 && 4.80 & 0.97 \\ \midrule
            Plain/ \\ Weird             & 2.26 & 0.90 && 4.88 & 0.98 \\ \midrule
            Ordinary/ \\ Uncanny        & 2.30 & 1.04 && 4.74 & 0.99 \\ \midrule
            \textbf{Average}        & 2.22 & 0.93 && 4.82 & 0.91 \\ 
        \bottomrule
    \end{tabular}
\caption{Mean and standard deviation of 7-point Likert scale score distributions per adjective pair (1-7), as well as the overall uncanniness score.  Wilcoxon-Signed-Rank tests~\cite{rey_wilcoxon-signed-rank_2011} for all attributes were significant at $p < 0.0002$.}
\label{table:uncanniness-results}
\end{table}

\begin{figure}[h]
    \centering
    \includegraphics[width=1\linewidth]{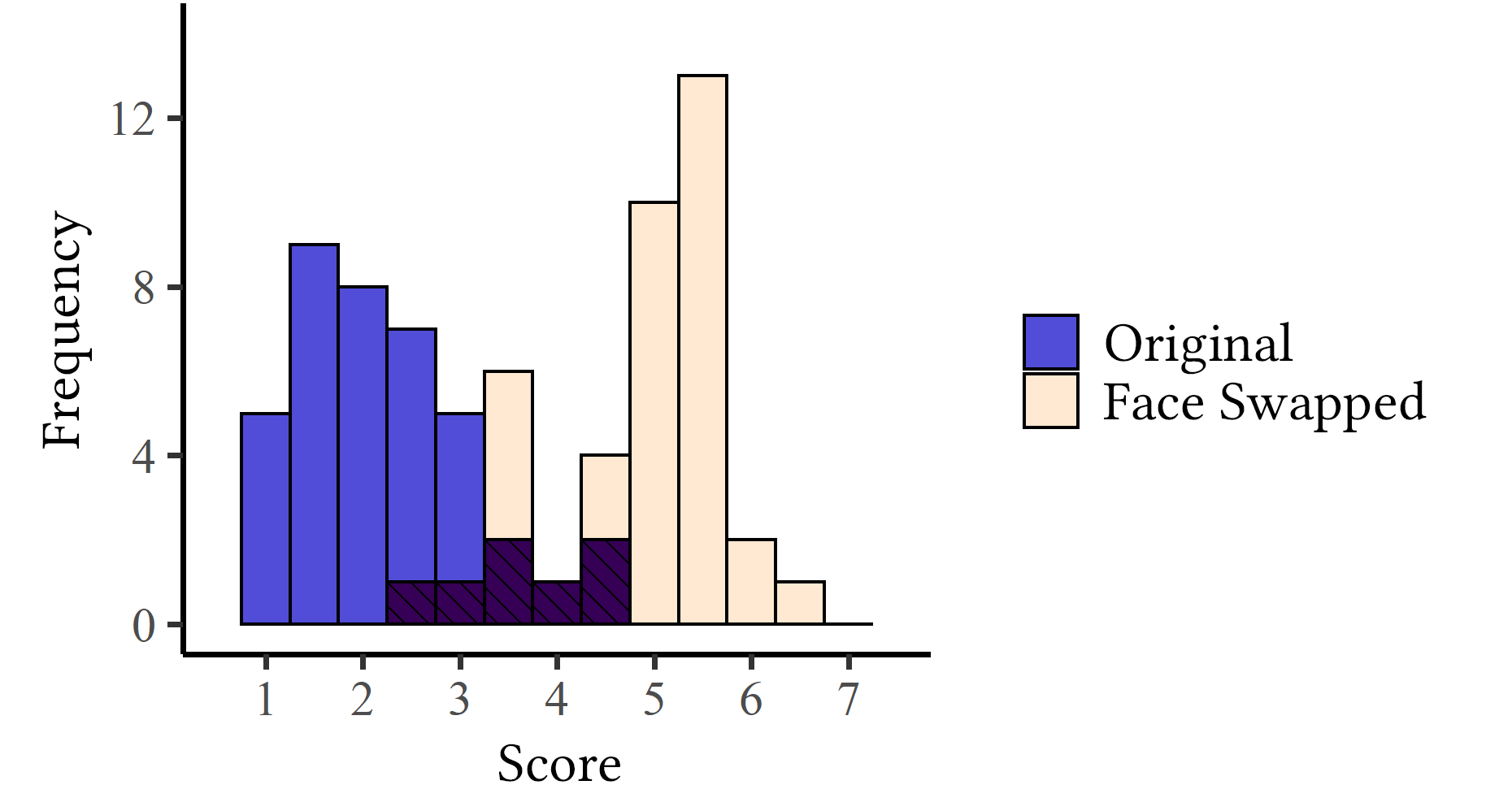}
    \caption{Histogram showing the distribution of participant responses between original videos and face swapped videos.  A higher score represents a higher overall uncanniness measurement.}
    \label{fig:uncanniness-hist}
\end{figure}

Participants' uncanniness measurements are averaged over the group of all twenty original videos and the group of all twenty face swapped videos, which results in each participant providing one overall measurement per group. The mean of the face swap measurements ($\mu=4.82, \sigma=0.91$) exceed the mean of the original measurements ($\mu=2.22, \sigma=0.93$). We perform the Wilcoxon-Signed-Rank test~\cite{rey_wilcoxon-signed-rank_2011} to test for the equality of these two groups.  Results show that the face swap and original uncanniness measurements are not equal and thus \textbf{rejects the null hypothesis for H1} ($p < 0.0002$).  Each adjective pair response, when analyzed independently, is significantly more negative for the face swapped video set when compared to the original.  Figure~\ref{fig:uncanniness-hist} shows the distribution of uncanniness measurements. The means and standard deviations broken down by adjective pair are additionally reported in Table~\ref{table:uncanniness-results}. 

\subsection{Comparison to Prior Analysis of Face Swaps}

\revision[In response to reviewer 2's comment, we have added a section comparing to Wohler et al.'s work and discussed how their limited analysis of uncanniness in a broader analysis of personality likely wasn't in-depth enough to see significance.]{In W\"{o}hler et al.'s work assessing the personality of face swaps as avatars~\cite{wohler_personality_2022}, both eerieness and viewer appeal are assessed.  These measurements could fall within the established uncanniness categories of \textit{eeriness} and \textit{attractiveness}~\cite{ho_measuring_2017}.  Their investigation did not find significant effects across a 7-way MANOVA, indicating similar personality ratings, and alluding to a lack of uncanniness of face swaps.  However, our work yields additional insights by focusing directly on uncanniness and discovering significant effects.  A number of factors could contribute to our findings not present in W\"{o}hler et al.'s~\cite{wohler_personality_2022}; for example, our set of bipolar adjective pairs probes more subtle viewer insights than a straight-forward prompt of \textit{"I found the person \{appealing, eerie\}}".  Additionally, there is more diversity in our swapped faces (where their dataset consists of young-adult German faces, FF++ DFD~\cite{rossler_faceforensics_2019} is a much more diverse set), and our stimuli containing multiple-individual conversations.  It could be that face swaps would be acceptable to users in suitable use cases, such as virtual avatars (with the proper permissions), but elicit uncanniness in cases where viewers must scrutinize whether the faces are real or fake.}
\section{Methodology}
\label{sec:methodology}

Considering our findings establishing that face swaps elicit uncanniness and previous work that establishes the eyes as a key region where viewer behavior differs~\cite{wohler_towards_2021, tahir_seeing_2021, gupta_eyes_2020}, we aim to improve the quality of generated gaze reconstructions in swapped faces.

We propose a novel method to explicitly prioritize gaze over all other implicitly defined facial attributes when training face swapping models.  The proposed method leverages a pretrained gaze estimation network, using the resulting gaze values to formulate a reconstruction loss focused on the eye region of the face.  Our approach is flexible and generalizes, meaning that it can be applied to any face swapping architecture and with other pretrained expert models.  Already-trained models could also be fine-tuned with this improvement.  We evaluate our method on DeepFaceLab (DFL)~\cite{liu_deepfacelab_2023}, comparing against both a gaze-unaware baseline model and their native solution, which had not been formally analyzed or explained in the literature.  \revision{To investigate improvements in gaze reconstruction from a quantitative standpoint,} we test the following hypotheses:

\begin{itemize}
    \item H2.1: Explicit gaze constraints during training will directly improve the accuracy of face swaps' gaze direction when compared to the original faces.
    \item H2.2: Models trained using gaze-centric loss terms derived from a pretrained gaze prediction model will have more accurate reconstructed gaze direction than models trained using pixel-based loss terms.
\end{itemize}

We further validate our results with a perceptual study.  We aim to assess how the proposed gaze constraints affect: (1.) users' ability to perform \textit{deepfake detection} (i.e., can users consciously tell that faces are real/ fake?), (2.) which regions of the face inform user decisions, and (3.) user responses regarding uncanniness.  We test the following hypotheses:

\begin{itemize}
    \item H3.1: Users are less capable of identifying face swaps that have been generated using models trained with explicit loss terms.
    \item H3.2: The eyes will be a less prominent factor in deepfake detection in face swaps generated using models trained with explicit loss terms.
    \item H3.3: Face swaps generated using models trained with explicit loss terms will be perceived as less uncanny by users than face swaps generated using baseline models.
\end{itemize}

\subsection{Overview of DeepFaceLab}

DFL is the most popular publicly available face swapping platform, so is representative of a large percentage of face swaps found online.  There are many resources online to aid in understanding DFL's pipeline\footnote{\url{https://mrdeepfakes.com/forums/thread-guide-deepfacelab-2-0-guide}}\footnote{\url{https://www.deepfakevfx.com/downloads/deepfacelab/}}. 
 For explanation and justification of DFL's model design, please refer to their publication~\cite{liu_deepfacelab_2023}.

\begin{figure*}[h]
    \centering
    \includegraphics[width=1\linewidth]{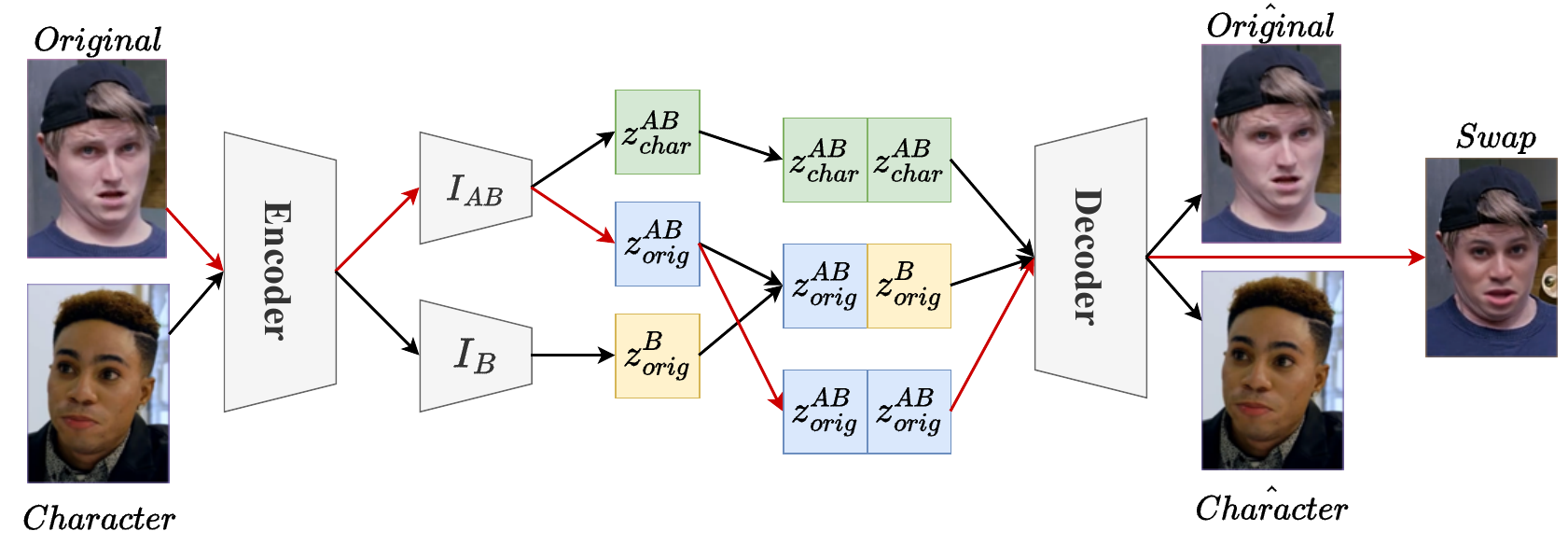}
    \caption{Illustration of DFL's LIAE architecture.  The pathway taken to create the resulting face swap is displayed in red.  Note that $z^{AB}_{char}$ is concatenated with a copy of itself to reconstruct the character face, and $z^{AB}_{orig}$ is concatenated with itself to produce the face swap result.}
    \label{fig:liae_design}
\end{figure*}

We use the Lightly Improved Auto-Encoder architecture (LIAE), which disentangles identity with intermediate networks between the encoder and decoder (see Figure~\ref{fig:liae_design}).  The first intermediate network $I_{AB}$ generates latent vectors $z^{AB}_{char}$ and $z^{AB}_{orig}$ during the training process.  The second intermediate network $I_B$ is only given the original identity to generate $z^{B}_{orig}$.  Before passing to the decoder, the latent vectors are concatenated: the original face's becomes $z^{AB}_{orig} || z^{B}_{orig}$ and the character face's concatenates a copy of itself to become $z^{AB}_{char} || z^{AB}_{char}$.  These latent vectors are passed through the respective decoders to reconstruct the input faces.  During face swapping, the original face is only passed through $I_{AB}$ and concatenated onto itself to generate latent code $z^{AB}_{orig} || z^{AB}_{orig}$, which is then fed through the decoder to generate a face swapped result.  

\revision[In response to reviewer 1's comments on figure 3 versus figure 4: The prior explanation did not make it clear that face swapping is implicitly learned, and that the training task is actually an image reconstruction problem.  The model's forked structure allows the network to essentially "hard-code" the identity information into the intermediate models.]{The training process of the LIAE architecture, and for the majority of paired face swapping algorithms, does not actually involve swapping of faces.  During training, the objective function only aims to optimize the reconstruction of the two input individuals' faces.  During training, $I_{AB}$ is tasked with the non-identity attributes of both faces plus the identity information of the character.  As a result, the identity information of the character face becomes baked into the weights of the $I_{AB}$ model.}  Intuitively, we can interpret the first latent vector $z_1$ to contain attributes, the second vector $z_2$ to contain identity information, and $z_1 || z_2$ to contain full facial information.  \revision{During face swapping, the $z^{AB}_{orig}$ is concatenated with itself, keeping non-identity attributes but superimposing the character identity which has been hardwired into the model.}
\removal{The latent vector $z^{AB}$ never represents the original face's identity during training, so becomes hardwired to the character face's identity.}  The LIAE design can be seen in Figure~\ref{fig:liae_design}. 

During training, DFL uses segmentation masks to isolate the error calculation to relevant parts of the face~\cite{bulat_how_2017}.  The three masks utilized are of the face ($M_{face}$), the eyes ($M_{eyes}$), and the eyes plus mouth ($M_{em}$).  In the following equations, we define the input faces as $Y$ and their reconstructions as $\hat{Y}$\footnote{Note that original and character faces' reconstruction loss are computed in identical fashion.}.  The reconstruction loss combines difference of structural similarity (DSSIM)~\cite{wang_image_2004, zhao_loss_2017} and mean squared error (MSE).  DSSIM enforces structural consistency between the input and output face using luminance, contrast, and structural components, and MSE error enforces pixel-wise similarity.  The loss equations are as follows:

{
\begin{gather}
    SSIM(Y, \hat{Y}) = \frac{(2\mu_i\mu_j + c_1)(2\sigma_{ij} + c_2)}{(\mu_i^2 + \mu_j^2 + c_1)(\sigma_i^2 + \sigma_j^2 + c_2)} \\ 
    L_{DSSIM}(Y, \hat{Y}) = \frac{1 - SSIM(Y, \hat{Y})}{2}
\end{gather}

\centering{where $i$, $j$ $=$ sliding windows of size $N$x$N$ 

$\mu_i$, $\mu_j$ $=$ average of $i$, $j$ $\qquad$ $\sigma_i^2$, $\sigma_j^2$ $=$ variance of $i$, $j$ $\qquad$ 

$\sigma_{ij}$ $=$ covariance of $i$, $j$ $\qquad$ $c_1$, $c_2$ $=$ stabilizing variables}

}

\begin{equation}
    L_{MSE}(Y, \hat{Y}) = \frac{1}{n} \sum_{i=1}^{n} (\hat{Y_i} - Y_i)^2
\end{equation}

The core reconstruction loss is a weighted sum between DSSIM, MSE, and an MSE calculation comparing the input and predicted face masks:

\begin{multline}
    \label{eqn:core_loss}
    L_{\triangle}(Y, \hat{Y}, M_{face}, \hat{M_{face}}) = \lambda_1 L_{DSSIM}(Y, \hat{Y}) + \\ 
    \lambda_2 L_{MSE}(Y, \hat{Y}) + \lambda_3 L_{MSE}(M_{face}, \hat{M_{face}})
\end{multline}

\revision[Here I have clarified the source of the DFL+em condition to address Reviewer 1's feedback.]{Within DFL's codebase, there is an optional \textit{eyes and mouth priority} parameter.  When enabled, an additional loss equation is applied which focuses on the eyes and the mouth (Equation \ref{eqn:em_loss}) by measuring the absolute value of pixel error between original and generated faces, masked to the eye and mouth regions.  Prior to our work, there existed no formal analysis of the effectiveness of this method, which is only mentioned briefly in DFL's associated publication~\cite{liu_deepfacelab_2023}.  We compare our proposed method both against DFL as a baseline, and DFL with this optional term enabled (which we refer to as DFL+em).  Thus, we also provide the first formal evaluation of the pixel-based approach to enforce gaze.}

\removal{DFL can explicitly target facial attributes via its optional eyes and mouth priority term.  This integrates well with the main loss equation, measuring the absolute value of pixel error between the original and generated faces masked to the eyes and mouth.  This is an optional term that must be enabled by DFL users.}

\begin{equation}
    \label{eqn:em_loss}
    L_{\triangle em}(Y, \hat{Y}, M_{em}) = \lambda_{em} |Y M_{em} - \hat{Y} M_{em}|
\end{equation}

\subsection{Proposed Gaze Reconstruction Loss}

Motivated by previous image generation methods' success using pretrained expert models, we leverage a gaze estimation network.  We incorporate L2CS-Net\footnote{https://github.com/Ahmednull/L2CS-Net}~\cite{abdelrahman_l2cs-net_2022}, which predicts pitch and yaw angles $\mu$, $\phi$ from input face images.  This network is optimized towards unconstrained environments so is well suited to the data typical in training face swaps.   We incentivize the face swapping model to better reconstruct gaze by penalizing offsets in predicted gaze angle between the input and reconstructed faces during training.

\revision[Reviewer 3 requested additional details regarding L2CS-Net.  We normalize the images by resizing and normalizing the mean and variance, following the implementation in their pipeline under "demo.py" for in-the-wild faces.  We do not perform aligning rotations to images as they did in their evaluations of benchmark datasets.]{Face images are upscaled and normalized by mean and variance before being processed by L2CS-Net.  The gaze angle vectors $\mu$, $\phi$ are returned for the real faces being analyzed, and $\hat{\mu}$, $\hat{\phi}$ for reconstructed images.  We compute the angular offset between these gaze vectors to then be used as a regularizing term during the face swap model's training.}

\begin{figure*}[t]
    \centering
    \includegraphics[width=1\linewidth]{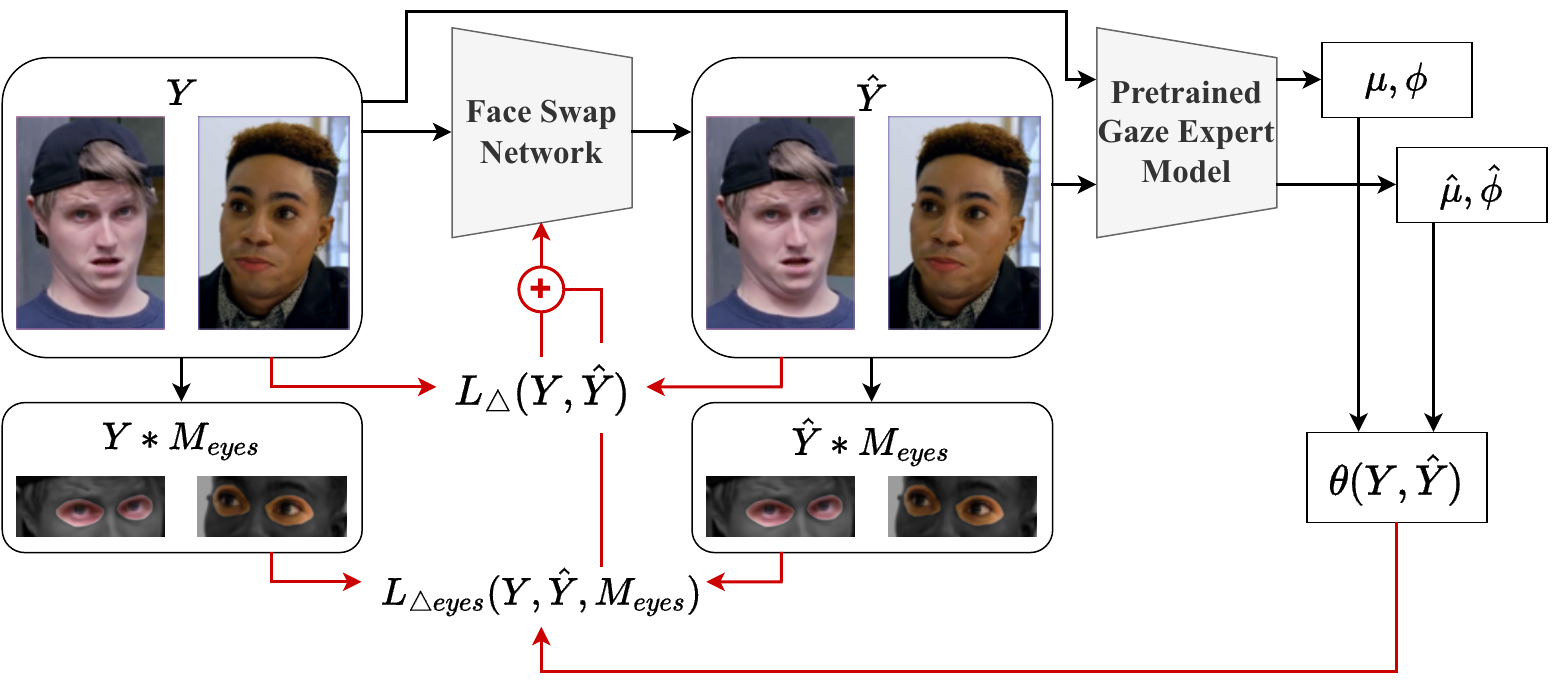}
    \caption{\revisionnote{With the clarifications in Section 4.1, it should now be clearer that the $Y$ faces are passed through the face swapping network to produce reconstructions $\hat{Y}$; during this training loop, face swapping does not actually occur.} Design diagram of the steps to compute the gaze reconstruction loss.}
    \label{fig:gaze_pipeline}
\end{figure*}

Our gaze reconstruction loss is computed as follows.  $\mu$ and $\phi$ are converted to normalized Cartesian coordinates, then the angle $\theta$ between the two vectors is found.
\begin{gather*}
    \mu, \phi = L2CS(Y) \qquad \qquad \hat{\mu}, \hat{\phi} = L2CS(\hat{Y}) \\
    x = sin(\phi) cos(\mu) \qquad \qquad y = sin(\phi) sin(\mu) \\ 
    z = cos(\phi) \qquad \qquad \qquad V_1 = <x, y, z> \\
    \hat{x} = sin(\hat{\phi}) cos(\hat{\mu}) \qquad \qquad \hat{y} = sin(\hat{\phi}) sin(\hat{\mu}) \\ 
    \hat{z} = cos(\hat{\phi}) \qquad \qquad \qquad V_2 = <\hat{x}, \hat{y}, \hat{z}>
\end{gather*}

\begin{equation}
    \label{eqn:theta_loss}
    \textrm{Error is computed as:} \qquad \theta(Y, \hat{Y}) = cos^{-1}\Biggl(\frac{V_1 \cdot V_2}{\|V_1\| \|V_2\|}\Biggr)
\end{equation}

We apply this error term only to the regions of the network that correspond to the eyes.  We use $Y$, $\hat{Y}$, and $M_{eyes}$ to construct a reconstruction loss specific to the eyes that can be scaled by the computed $\theta$ and hyperparameters $\alpha$ and $\beta$.  We structure our loss equation similarly to equation~\ref{eqn:core_loss}, using DSSIM and MSE computations on the original and reconstructed image eye regions.

\begin{multline}
    \label{eqn:gaze_loss}
    L_{\triangle eyes}(Y, \hat{Y}, M_{eyes}) = \theta(Y, \hat{Y}) \Bigl(\alpha L_{DSSIM}(Y M_{eyes}, \hat{Y} M_{eyes}) + \\
    \beta L_{MSE}(Y M_{eyes}, \hat{Y} M_{eyes})\Bigr)
\end{multline}

An illustration of the design and steps taken to compute the loss equation can be seen in Figure~\ref{fig:gaze_pipeline}.

The proposed method could alternatively be enabled later on in training to fine-tune the network, opening the possibility of enhancing existing models.  We investigate this in our analysis, pretraining the network with the baseline method then enabling the gaze reconstruction loss during the final phase of training.

\subsection{Evaluation Dataset}
\label{sec:evaluation_dataset}

We generate a small dataset to serve as an evaluation platform for the proposed approach.  We generate our face swaps using the source video clips taken from the FF++ DFD dataset~\cite{rossler_faceforensics_2019}.  In the dataset, subjects perform the same tasks\footnote{The video segments we use are: exit phone room, kitchen pan, outside talking pan laughing, walking outside cafe disgusted.  These are concatenated into a single video $\sim$2 minutes in length per subject.}, ensuring similar expression and head pose, making these clips ideal for high quality face swaps.  Our dataset consists of 6 subjects (3 female, 3 male).  For each gender, two subjects have similar appearance to one another.  Per gender, we permute all combinations of subjects being used as the character and original face, resulting in a total of 12 unique face pairs, 6 per gender. 

\begin{figure*}[h]
    \centering
    \includegraphics[width=1\linewidth]{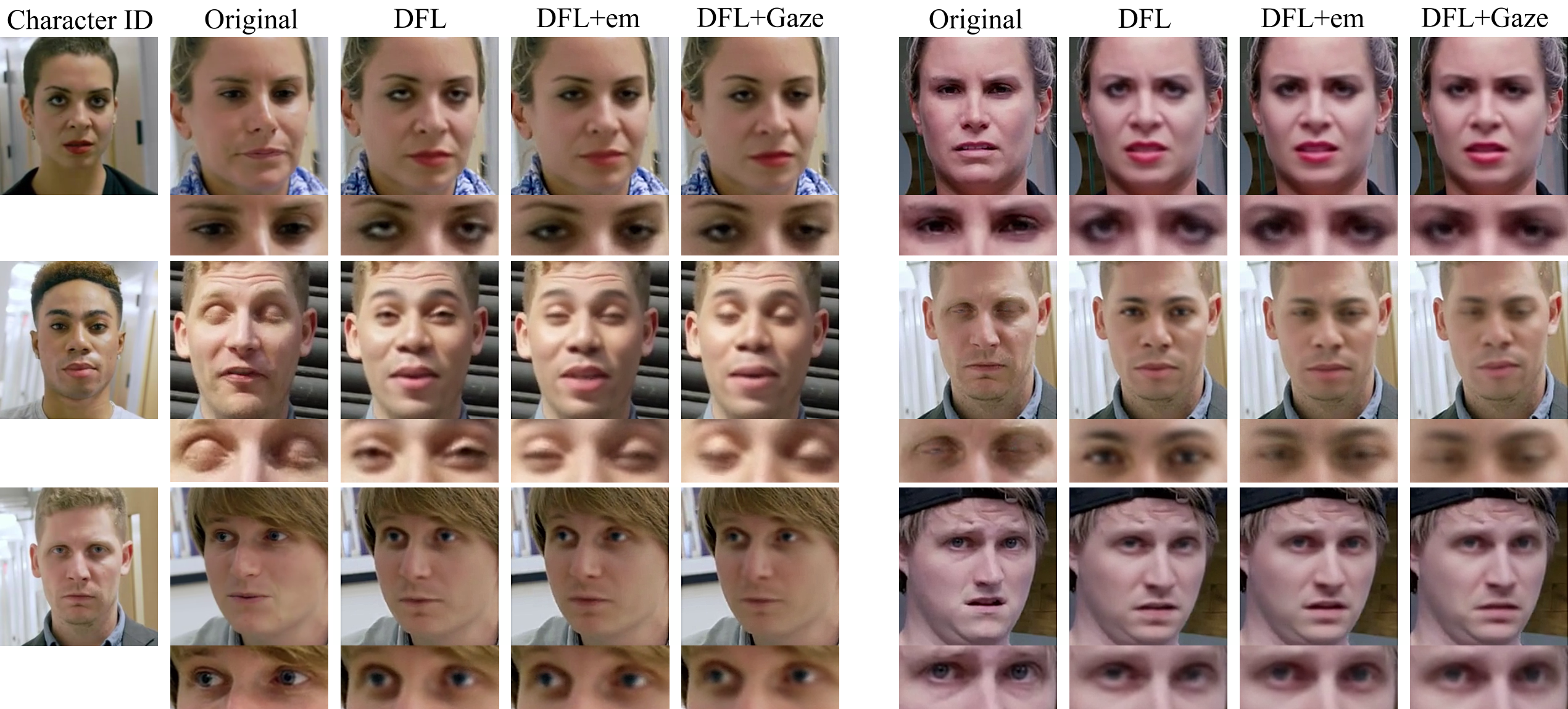}
    \caption{\revisionnote{Doubled the number of exemplars.} Visual comparison of face swaps produced by the baseline DFL method, DFL with eyes and mouth priority loss (em), and DFL with our proposed loss (gaze).  Both improvements over the baseline reduce gaze angle error.}
    \label{fig:gaze_improvements}
\end{figure*}

We generate face swaps across multiple conditions, keeping all other hyperparameters consistent.  All faces are generated at 128x128 resolution.  Every model is pretrained for 100 thousand iterations on the CelebA dataset~\cite{liu_deep_2015}, then trained for the final 20 thousand iterations on the identity pair.  \revision{In all conditions, only the training phase is altered and all other components (face detection, normalization, video generation, etc.) are achieved with DFL's pipeline.}  Frames from our generated dataset can be seen in Figure~\ref{fig:gaze_improvements}.  The conditions are:

\begin{itemize}
    \item \textbf{DFL.}  The model implicitly learns gaze behavior while optimizing the core reconstruction loss in Equation~\ref{eqn:core_loss}.

    \item \textbf{DFL+em.}  DeepFaceLab with eyes and mouth priority loss enabled (see Equation~\ref{eqn:em_loss}).  DFL's native solution which further enforces pixel-wise similarity for the key regions of the face.

    \item \textbf{DFL+Gaze.}  DeepFaceLab with our proposed gaze loss (see Equations~\ref{eqn:theta_loss} \&~\ref{eqn:gaze_loss}).  The model explicitly enforces consistency using gaze vectors computed by the pretrained expert model.

    \item \textbf{DFL+Gaze (finetuning).}  The model is pretrained with no gaze-specific loss, then trained for the final 20 thousand iterations using our proposed loss.

    \item \textbf{DFL+em+Gaze.}  Both \removal{DFL's native approach}\revision{em} and our proposed approach are enabled during training.
\end{itemize}

\subsection{Perceptual Validation Survey}

Our perceptual study provides further evidence towards the effectiveness of gaze-centric loss terms when training face swapping models.  These results give insights beyond quantitative error measurements.  We implement a between-subjects design where participants view videos with the same content, but generated using different face swapping conditions.  Our independent variables are four generation conditions: DFL, DFL+em, DFL+Gaze, and real (unmodified videos of real faces).  Our dependent variables are scorings for deepfake detection, prevalence of attributes selected as aiding in deepfake detection, and uncanniness measurement scores.

\subsubsection{Stimuli} 

The videos presented are a subset of the evaluation dataset described in Section~\ref{sec:evaluation_dataset}.  20 total videos are presented, with 18 being face swaps and 2 being videos of real individuals brought in from FF++ DFD to serve as a control condition.  The video clips are trimmed to talking segments.  In every video, the subject of the clip is first shown facing the left and talking to another person facing opposite the camera.  Then, the same subject is shown in another scene directly facing the camera and talking\footnote{These talking clips correspond to the kitchen pan and outside talking pan laughing clips in FF++ DFD.}.  \revision{We preserve the entirety of the selected video segments, in which the actors perform the same actions, rather than trimming to a number of frames.}  The resulting video clips are between 30 seconds and 1 minute in length.

For each gender, we select 3 face-body pairs such that each face and each body is represented once.  An additional video is brought in directly from the FF++ DFD dataset with no face swap manipulation applied.  To provide consistent resolution with other stimuli, the faces of the control videos are extracted, resized to 128x128, then replaced in each video frame.  This results in 6 swapped face-body pairs and 2 real individuals.  For each of the swapped face-body pairs, we generate 3 videos with models trained under the 3 conditions: DFL, DFL+em, and DFL+Gaze.  

\subsubsection{Participants} 

Survey participants were recruited under IRB approved protocol via several communication channels including word of mouth and electronic mailing list advertisements ($N=109$; 51.38\% male, 47.71\% female, 0.92\% other).  The survey population consists mainly of undergraduate University students. The racial-ethnic distribution is 63.30\% White, 26.61\% Asian, 4.59\% Black or African American, 13.76\% Hispanic/Latino, where 12.84\% of participants report two or more races.  The median age is 20 years (IQR = 20-26). Survey data is anonymized for subsequent analysis.

\subsubsection{Procedure}

The survey was conducted online and hosted via Qualtrics.  Prior to taking the survey, participants were given a definition of face swapping and informed that the videos seen may or may not be face swaps.  Participants were then shown one stimulus video at a time and were asked to watch each video in its entirety.  

The study consists of a deepfake detection task paired with an uncanniness survey.  For each video seen, participants were instructed to move a slider indicating the likelihood that they believed the shown video to be a face swap.  A score of 0 corresponds to absolute certainty that the video was real, and a score of 100 corresponds to absolute certainty that the video contained a face swap.  Participants were also asked to denote which facial attributes affect their decision, utilizing the subset of attributes identified by~\cite{tahir_seeing_2021} pertaining to the face.  Participants were then asked to select all facial attributes that helped to make their decision.  The choices are \textit{forehead/hair}, \textit{nose}, \textit{chin/jaw}, \textit{eyes}, \textit{mouth}, \textit{eyebrows}, \textit{cheeks}, \textit{other}, and \textit{none}.  We asked participants to denote these key attributes regardless of their decision of real or fake; if participants falsely identify that videos are real but choose an attribute as aiding their decision, this indicates that the attribute mimicked natural behavior at a high level.  Participants were finally asked to rate the stimulus on bipolar adjective Likert scales as defined in Section~\ref{sec:uncanniness_protocol} to measure uncanniness. 

Each participant saw every face-body pair once and saw each face swap condition once per gender.  The distribution of conditions and face-body pairs evaluated are balanced across participants.  To further explain, one participant would see the pair \{face 1, body 2\} generated with the DFL model, but another participant would see the same face-body pair generated with the DFL+Gaze model instead.  All participants see all videos of real individuals.  On average, the survey would take 15-20 minutes.
\section{Results}
\label{sec:results}

We first evaluate our generated dataset quantitatively.  The dataset consists of 12 face-body pairs, with a separate model trained for each condition, yielding $N=60$ models and videos in total.  We then present the results of our perceptual study, providing qualitative evidence towards the efficacy of the evaluated gaze loss terms.

\subsection{Quantitative Evaluation}

We assess the performance of each gaze loss condition by analyzing the offset in viewing angles between resulting face swaps and the real faces across the corresponding source video.  To compute this metric, we utilize L2CS-Net~\cite{abdelrahman_l2cs-net_2022} to predict a gaze viewing angle for each condition, considering the source video's predicted gaze vector to be the ground truth.  In our evaluation we use DFL's internal parameters for our $\lambda$ values.  Namely, $\lambda_1, \lambda_2, \lambda_3 = 10$, $\lambda_em = 300$.  When implementing our proposed loss term, we use $\alpha = 3$ and $\beta = 30$.

We analyze error values, collapsing from individual frames ($\sim$2900 per video) to average across each individual in the dataset.  The baseline DFL produces an average error of 5.98\textdegree \space[95\% Confidence Interval (CI): 4.82, 7.13].  All improvements on the baseline method produce noticeably more accurate gaze values:  DFL+em averages 4.85\textdegree \space[95\% CI: 3.80, 5.90], DFL+Gaze averages 4.71\textdegree \space[95\% CI: 3.66, 5.77], DFL+Gaze (finetuning) averages 4.85\textdegree \space[95\% CI: 3.80, 5.90], and DFL+em+Gaze averages 4.72\textdegree \space[95\% CI: 3.67, 5.77].  On the test dataset, introducing DFL's eyes and mouth priority term decreases reconstructed gaze error by 18.9\%; introducing the proposed method decreases by 21.2\%, and introducing both components decreases gaze error by 21.1\%.

\begin{figure}[h]
    \centering
    \includegraphics[width=1\linewidth]{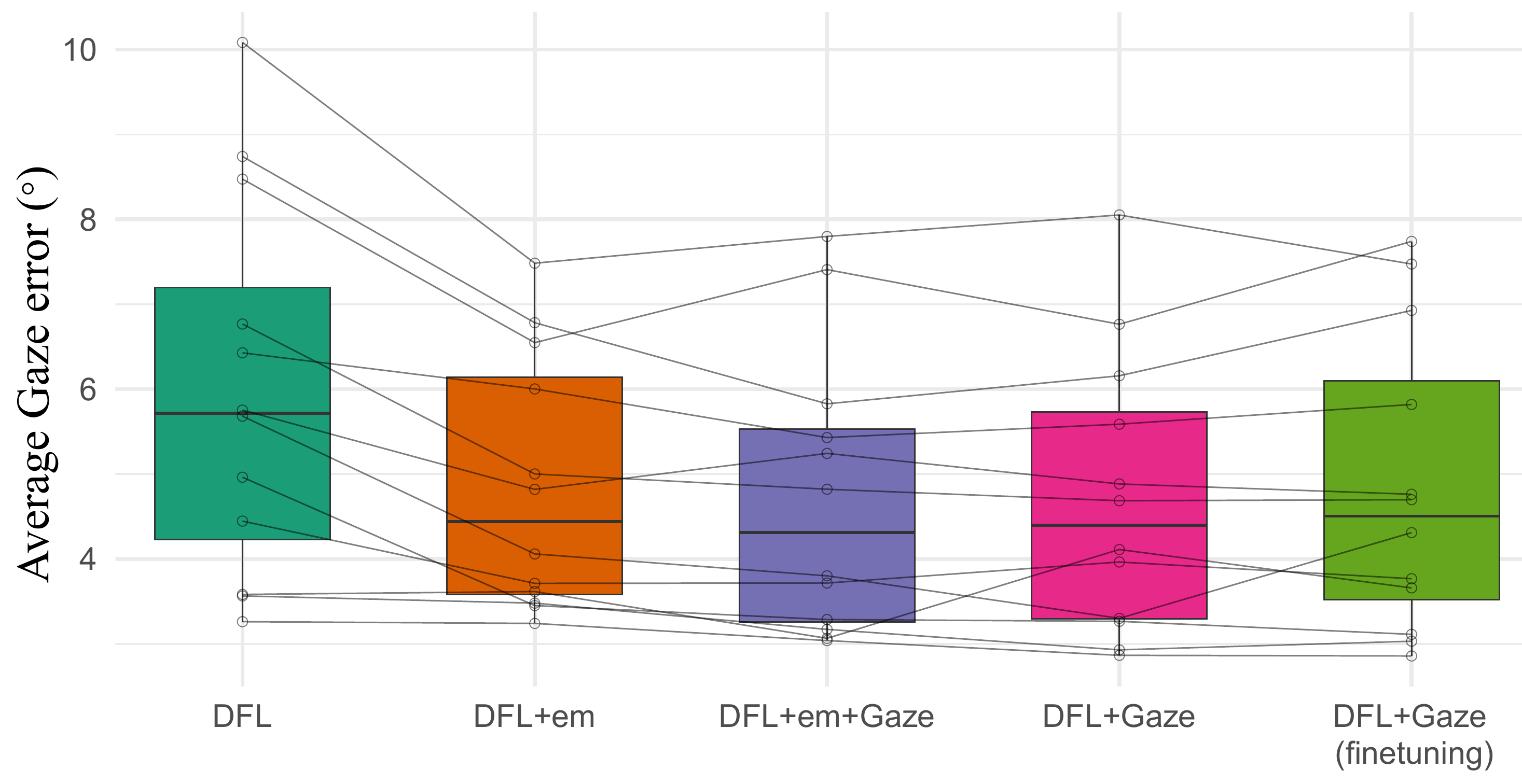}
    \caption{Plot of mean gaze error across all evaluated videos ($N=12$) by condition.  Individual video results are plotted over-top and connected across each box plot.}
    \label{fig:box_whisker}
\end{figure}

We test for significance via a linear mixed-effects model.  We first compute the average of the log of angular error for each method and individual, applying the log transform to improve normality of error distributions. We then model errors as $average(log($error$))$ $\sim$ method with a random intercept per individual.  All evaluated methods significantly improve over the baseline DFL ($p < 0.001$), \textbf{supporting hypothesis H2.1}.  However, we have not found statistical evidence pair-wise between any of the improved methods, and are \textbf{unable to reject the null hypothesis for H2.2}.  Interestingly, the DFL+em+Gaze approach combining pixel information and explicit gaze modeling yielded insignificant benefit over DFL+em (t$(1,44)=1.603, p=0.116$).  This may indicate that the two optimizations capture similar underlying information.

Each method's performance across individuals in the dataset is plotted in Figure~\ref{fig:box_whisker}.  We see a large amount of variability among individual video results in all methods other than DFL, indicating roughly equivalent performance for all improvements analyzed.  


\subsection{Perceptual Evaluation}

We evaluate users' perceptual feedback across the four evaluated conditions: real, DFL, DFL+em, DFL+Gaze.  We present our results across 3 metrics of analysis: deepfake detection results, attribute importance, and uncanniness.  Note that the study is a between-subjects design.  While every viewer witnesses the 2 real videos, each viewer witnesses 6 face swapped videos under the DFL, DFL+em, and DFL+Gaze conditions, uniformly distributed amongst participants.  We are able to directly compare between the face swap conditions to measure the minute differences produced by each face swap across all viewers; however, the real videos are distinct and serve mainly as a control condition.

\subsubsection{Deepfake Detection}
\label{sec:deepfake-detection}

\begin{figure}[h]
    \centering
    \includegraphics[width=0.8\linewidth]{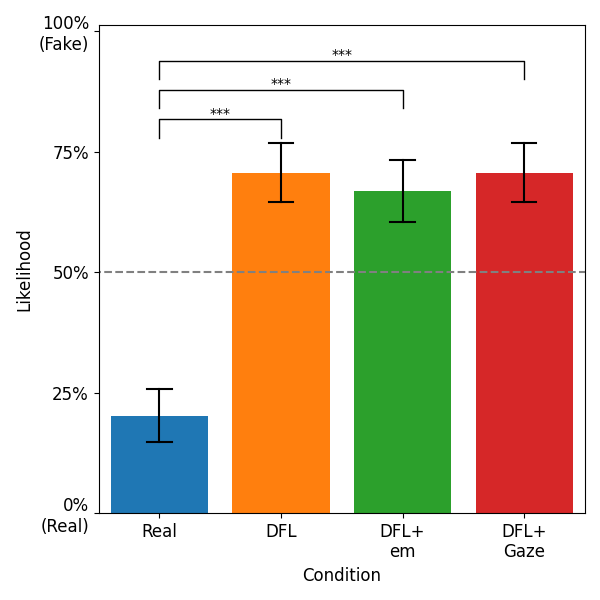}
    \caption{Coded user responses for the deepfake detection task.  Viewer responses ranging from [0-100] are remapped to 0 and 1.  \revisionnote{This line was accidentally left in from a previous analysis where we mapped to three values rather than binary.  We have verified that this was the only inconsistency, and that the in-text values are correct.} \removal{However, responses are mapped to 0.5 if originally between 40-60\%, reflecting that viewers were not confident in their decision.}  Significant differences are reported for the Mann-Whitney U test~\cite{mcknight_mann-whitney_2010}.  95\% CIs are displayed.}
    \label{fig:likelihood-bar}
\end{figure}

We first analyze viewers' ability to detect whether videos are real or fake under the different face swap conditions analyzed.  Viewer responses are confidence values that the viewed video is a deepfake on a continuous scale from 0 to 100\%.  Following the deepfake detection methodology from~\cite{groh_deepfake_2022}, we code viewer confidence values into concrete classifications of real or fake, modeling average performance across the population.  \revision[Based on reviewer 1's feedback, we have elaborated on the justification (originally from Groh et al.) to code confidence values into a binary choice.]{Through this presentation to participants, viewers tend to more deeply analyze the presented faces while considering their own confidence, rather than presenting participants with a two-alternative forced choice.}  Viewer responses \textit{below} 50\% are coded as 0, indicating that the viewer believed the video to be real.  Reponses \textit{above} 50\% are coded as 1, indicating that the viewer believed the video to be a fake.  By averaging across all viewer responses, we receive the average likelihood that a viewer would believe the video to be a fake.

For the real video stimuli, the likelihood is 20.19\%  [95\% CI: 14.74, 25.65].  For baseline DFL, the likelihood is 70.67\% [95\% CI: 64.48, 76.86].  DFL with pixel-based loss (DFL+em) is 66.83\% [95\% CI: 60.43, 73.22] and DFL with the proposed gaze estimation loss (DFL+Gaze) is 70.67\% [95\% CI: 64.48, 76.86].  While all face swap conditions are statistically significant ($p < 0.001$) via the Mann-Whitney U test~\cite{mcknight_mann-whitney_2010} compared to the real video condition; we have not found statistical significance when comparing DFL+em and DFL+Gaze against the baseline DFL.  Figure~\ref{fig:likelihood-bar} displays the average likelihood for each stimuli across all participants.  Based on these findings, \textbf{we can not confirm H3.1}, as we do not see significant changes in deepfake detection between conditions.

\subsubsection{Attribute Importance}

When participants perform the deepfake detection task, they are also prompted to select which facial attributes aid in their decision.  This choice provides information both when viewers correctly perform deepfake detection and when incorrect.  If a specific attribute is chosen to aid the viewer in correctly identifying a video as fake, then the attribute is a \textit{telling factor} which is likely not correctly reconstructed from the source video.  Alternatively, if the attribute is chosen as aiding the viewer in \textbf{incorrectly} labeling a fake video as real, then the attribute functions as a \textit{red herring}, being realistic enough to trick the viewer.

We report the prevalence of each attribute as how often it is selected as a factor in the videos that participants perceived as either real or as deepfakes, respectively. We additionally report the sum prevalence across all videos regardless of prediction.

We report these metrics for all attributes in Table~\ref{tab:attributes-table}.  As we are particularly focused on the eyes, we report the prevalence of the eyes for our face swap conditions here.  Across the videos predicted as deepfakes (the viewer has correctly identified the video as fake), DFL prevalence is 71.23\% \revision{[95\% CI: 63.89, 78.58]}, DLF+em prevalence is 67.41\% \revision{[95\% CI: 59.5, 75.31]}, and DFL+Gaze prevalence is 62.94\% \revision{[95\% CI: 55.02, 70.85]}.  These results confirm that the eyes are a key factor in human deepfake detection, being chosen substantially more often as a feature than other attributes.  Additionally, both improvements over the baseline DFL result in a reduced prevalence of the eyes as a factor.  \revision[The below sections were accidentally misleading as we had forgotten to include statistical test results.  Our proposed approach is significant against the baseline, whereas DFL+em is not, yet we do not find statistical significance between the two.  The prior wording could have been interpreted as DFL+Gaze is significantly better than DFL+em.]{We verify significance across conditions using the Kruskal-Wallis H test ($H = 16.27$, $p < 0.001$).  We investigated individual effect via the Mann-Whitney U Test~\cite{mcknight_mann-whitney_2010} between each pair of conditions.  All conditions are significant against real videos ($p < 0.001$), but further measures against the DFL baseline are not shown to be significant.  DFL+Gaze has a reduced prevalence of the eyes of 8.29\% compared to the baseline DFL ($p = 0.13$), whereas DFL+em reduced eye prevalence by only 3.82\% ($p = 0.48$).  Interaction between DFL+Gaze and DFL+em is not significant ($p = 0.43$)}.  These results \textbf{provide evidence towards H3.2}.  However, across all face swap methods the eyes are selected over 50\% of the time when making the \textit{incorrect} decision.  This likely indicates that while the eyes can be a key feature for deepfake detection, it is easy for individuals to attribute a false sense of security in their choice by focusing on the eyes.

The mouth is also an attribute of interest, and the second most prominent attribute selected to aid deepfake detection.  Results for the mouth are generally close to 50\% when correctly or incorrectly classifying the face swap videos, indicating that the mouth may not be a reliable feature despite viewers considering it perceptually important.  However, it is important to note that the DFL+em condition places pixel loss on a masked region of both the eyes and mouth, while the DFL+Gaze condition focuses only on the eyes.  We see DFL+em decrease the presence of the mouth during correct deepfake prediction by 9.72\%, while DFL+Gaze increases presence by 13.01\%.  This indicates that there is a trade-off present; by solely focusing on the eyes, the accuracy and realism of the generated mouth may decrease.

\begin{table}[th!]
\small
\centering
\setlength{\extrarowheight}{0pt}
\addtolength{\extrarowheight}{\aboverulesep}
\addtolength{\extrarowheight}{\belowrulesep}
\setlength{\aboverulesep}{0pt}
\setlength{\belowrulesep}{0pt}
\caption{User responses to attributes selected as aiding in their deepfake detection decision.  The columns \textit{Real} and \textit{Deepfake} indicate participant selections when the stimuli was \textit{perceived} as the respective category.  Cells in which users correctly identified the true class of the video are highlighted.}
\label{tab:attributes-table}
\begin{tabular}{cl|ccc} 
\toprule
\multicolumn{1}{l}{}                                                                                 & \multicolumn{1}{l}{}                 & \multicolumn{3}{c}{\textbf{Influence on user predictions }}                                                 \\
\multicolumn{1}{l}{\textbf{Attribute}}                                                               & \multicolumn{1}{l}{\textbf{Stimuli}} & \textbf{Real}                               & \textbf{Deepfake}                           & \textbf{Total}  \\ 
\hline
\multirow{4}{*}{\rotcell{\begin{tabular}[c]{@{}c@{}}\textbf{Forehead/}\\\textbf{Hair}\end{tabular}}} & \textbf{Real}                        & {\cellcolor[rgb]{0.922,0.922,0.922}}47.62\% & 22.50\%                                     & 42.79\%         \\
                                                                                                     & \textbf{DFL}                         & 38.71\%                                     & {\cellcolor[rgb]{0.922,0.922,0.922}}31.51\% & 33.65\%         \\
                                                                                                     & \textbf{DFL+em}                      & 41.10\%                                     & {\cellcolor[rgb]{0.922,0.922,0.922}}42.22\% & 41.83\%         \\
                                                                                                     & \textbf{DFL+Gaze}                    & 43.08\%                                     & {\cellcolor[rgb]{0.922,0.922,0.922}}37.06\% & 38.94\%         \\ 
\hline
\multirow{4}{*}{\rotcell{\textbf{Eyes}}}                                                             & \textbf{Real}                        & {\cellcolor[rgb]{0.922,0.922,0.922}}67.86\% & 37.50\%                                     & 62.02\%         \\
                                                                                                     & \textbf{DFL}                         & 59.68\%                                     & {\cellcolor[rgb]{0.922,0.922,0.922}}71.23\% & 67.79\%         \\
                                                                                                     & \textbf{DFL+em}                      & 56.16\%                                     & {\cellcolor[rgb]{0.922,0.922,0.922}}67.41\% & 63.46\%         \\
                                                                                                     & \textbf{DFL+Gaze}                    & 60.00\%                                     & {\cellcolor[rgb]{0.922,0.922,0.922}}62.94\% & 62.02\%         \\ 
\hline
\multirow{4}{*}{\rotcell{\textbf{Eyebrows}}}                                                         & \textbf{Real}                        & {\cellcolor[rgb]{0.922,0.922,0.922}}26.79\% & 30.00\%                                     & 27.40\%         \\
                                                                                                     & \textbf{DFL}                         & 17.74\%                                     & {\cellcolor[rgb]{0.922,0.922,0.922}}19.18\% & 18.75\%         \\
                                                                                                     & \textbf{DFL+em}                      & 15.07\%                                     & {\cellcolor[rgb]{0.922,0.922,0.922}}22.96\% & 20.19\%         \\
                                                                                                     & \textbf{DFL+Gaze}                    & 21.54\%                                     & {\cellcolor[rgb]{0.922,0.922,0.922}}23.78\% & 23.08\%         \\ 
\hline
\multirow{4}{*}{\rotcell{\textbf{Nose}}}                                                             & \textbf{Real}                        & {\cellcolor[rgb]{0.922,0.922,0.922}}13.69\% & 7.50\%                                      & 12.50\%         \\
                                                                                                     & \textbf{DFL}                         & 17.74\%                                     & {\cellcolor[rgb]{0.922,0.922,0.922}}11.64\% & 13.46\%         \\
                                                                                                     & \textbf{DFL+em}                      & 15.07\%                                     & {\cellcolor[rgb]{0.922,0.922,0.922}}19.26\% & 17.79\%         \\
                                                                                                     & \textbf{DFL+Gaze}                    & 10.77\%                                     & {\cellcolor[rgb]{0.922,0.922,0.922}}20.28\% & 17.31\%         \\ 
\hline
\multirow{4}{*}{\rotcell{\textbf{Mouth}}}                                                            & \textbf{Real}                        & {\cellcolor[rgb]{0.922,0.922,0.922}}64.88\% & 62.50\%                                     & 64.42\%         \\
                                                                                                     & \textbf{DFL}                         & 59.68\%                                     & {\cellcolor[rgb]{0.922,0.922,0.922}}53.42\% & 55.29\%         \\
                                                                                                     & \textbf{DFL+em}                      & 46.58\%                                     & {\cellcolor[rgb]{0.922,0.922,0.922}}43.70\% & 44.71\%         \\
                                                                                                     & \textbf{DFL+Gaze}                    & 47.69\%                                     & {\cellcolor[rgb]{0.922,0.922,0.922}}66.43\% & 60.58\%         \\ 
\hline
\multirow{4}{*}{\rotcell{\textbf{Cheeks}}}                                                           & \textbf{Real}                        & {\cellcolor[rgb]{0.922,0.922,0.922}}25.00\% & 12.50\%                                     & 22.60\%         \\
                                                                                                     & \textbf{DFL}                         & 22.58\%                                     & {\cellcolor[rgb]{0.922,0.922,0.922}}21.23\% & 21.63\%         \\
                                                                                                     & \textbf{DFL+em}                      & 41.10\%                                     & {\cellcolor[rgb]{0.922,0.922,0.922}}23.70\% & 29.81\%         \\
                                                                                                     & \textbf{DFL+Gaze}                    & 24.62\%                                     & {\cellcolor[rgb]{0.922,0.922,0.922}}25.87\% & 25.48\%         \\ 
\hline
\multirow{4}{*}{\rotcell{\begin{tabular}[c]{@{}c@{}}\textbf{Chin/}\\\textbf{Jaw}\end{tabular}}}      & \textbf{Real}                        & {\cellcolor[rgb]{0.922,0.922,0.922}}19.64\% & 42.50\%                                     & 24.04\%         \\
                                                                                                     & \textbf{DFL}                         & 27.42\%                                     & {\cellcolor[rgb]{0.922,0.922,0.922}}21.92\% & 23.56\%         \\
                                                                                                     & \textbf{DFL+em}                      & 31.51\%                                     & {\cellcolor[rgb]{0.922,0.922,0.922}}25.19\% & 27.40\%         \\
                                                                                                     & \textbf{DFL+Gaze}                    & 20.00\%                                     & {\cellcolor[rgb]{0.922,0.922,0.922}}24.48\% & 23.08\%         \\ 
\hline
\multirow{4}{*}{\rotcell{\textbf{Other}}}                                                            & \textbf{Real}                        & {\cellcolor[rgb]{0.922,0.922,0.922}}16.67\% & 12.50\%                                     & 15.87\%         \\
                                                                                                     & \textbf{DFL}                         & 12.90\%                                     & {\cellcolor[rgb]{0.922,0.922,0.922}}17.12\% & 15.87\%         \\
                                                                                                     & \textbf{DFL+em}                      & 15.07\%                                     & {\cellcolor[rgb]{0.922,0.922,0.922}}14.81\% & 14.90\%         \\
                                                                                                     & \textbf{DFL+Gaze}                    & 13.85\%                                     & {\cellcolor[rgb]{0.922,0.922,0.922}}14.69\% & 14.42\%         \\ 
\hline
\multirow{4}{*}{\rotcell{\textbf{None}}}                                                             & \textbf{Real}                        & {\cellcolor[rgb]{0.922,0.922,0.922}}3.57\%  & 5.00\%                                      & 3.85\%          \\
                                                                                                     & \textbf{DFL}                         & 9.68\%                                      & {\cellcolor[rgb]{0.922,0.922,0.922}}0.68\%  & 3.37\%          \\
                                                                                                     & \textbf{DFL+em}                      & 15.07\%                                     & {\cellcolor[rgb]{0.922,0.922,0.922}}0.74\%  & 5.77\%          \\
                                                                                                     & \textbf{DFL+Gaze}                    & 10.77\%                                     & {\cellcolor[rgb]{0.922,0.922,0.922}}2.10\%  & 4.81\%          \\
\bottomrule
\end{tabular}
\end{table}

\subsubsection{Uncanniness}

\begin{table}[th!]
\centering
\caption{Mean and standard deviation of 7-point Likert scale responses per adjective pair across each of the evaluated stimuli.  All face swap conditions are statistically significant against the \textit{Real} videos via the Mann-Whitney U test~\cite{mcknight_mann-whitney_2010}.}
\begin{tblr}{
  cell{2}{2} = {r},
  cell{2}{3} = {r},
  cell{2}{4} = {r},
  cell{2}{5} = {r},
  cell{3}{2} = {r},
  cell{3}{3} = {r},
  cell{3}{4} = {r},
  cell{3}{5} = {r},
  cell{4}{2} = {r},
  cell{4}{3} = {r},
  cell{4}{4} = {r},
  cell{4}{5} = {r},
  cell{5}{2} = {r},
  cell{5}{3} = {r},
  cell{5}{4} = {r},
  cell{5}{5} = {r},
  cell{6}{2} = {r},
  cell{6}{3} = {r},
  cell{6}{4} = {r},
  cell{6}{5} = {r},
  cell{7}{2} = {r},
  cell{7}{3} = {r},
  cell{7}{4} = {r},
  cell{7}{5} = {r},
  vline{2} = {-}{},
  hline{1,8} = {-}{0.08em},
  hline{2,3,4,5,6,7} = {-}{},
}
{\textbf{Attribute}\\\textbf{Pair}} & \textbf{Real} & \textbf{DFL}  & {\textbf{DFL+}\\\textbf{em}} & {\textbf{DFL+}\\\textbf{Gaze}} \\
{Real/\\Synthetic}                  & {2.17\\$\pm$1.32} & {4.37\\$\pm$1.84} & {4.11\\$\pm$1.91}                & {4.35\\$\pm$1.95}                  \\
{Agreeable/\\Repulsive}             & {2.23\\$\pm$1.25} & {3.48\\$\pm$1.48} & {3.25\\$\pm$1.49}                & {3.49\\$\pm$1.58}                  \\
{Unremarkable/\\Unusual}            & {2.68\\$\pm$1.44} & {4.18\\$\pm$1.53} & {4.01\\$\pm$1.65}                & {4.10\\$\pm$1.68}                  \\
{Plain/\\Weird}                     & {2.59\\$\pm$1.35} & {4.14\\$\pm$1.65} & {3.92\\$\pm$1.67}                & {4.09\\$\pm$1.74}                  \\
{Ordinary/\\Uncanny}                & {2.40\\$\pm$1.45} & {4.23\\$\pm$1.74} & {3.96\\$\pm$1.75}                & {4.06\\$\pm$1.87}                  \\
\textbf{Average}                    & {2.41\\$\pm$1.17} & {4.08\\$\pm$1.42} & {3.85\\$\pm$1.48}                & {4.02\\$\pm$1.57}                  
\end{tblr}
\label{tab:perceptual-uncanniness}
\end{table}

Average bipolar adjective scores and standard deviations are reported in Table~\ref{tab:perceptual-uncanniness}.  The average score for the real videos is 2.41$\pm$1.17.  The average uncanniness for the baseline face swap condition is 4.08$\pm$1.42; DFL+em is 3.85$\pm$1.48; DFL+Gaze is 4.02$\pm$1.57.  The results cover a similar distribution to uncanniness metrics collected on the FF++ DFD dataset in Table~\ref{table:uncanniness-results}.  Despite the trends seen in the deepfake detection task, we are unable to derive significant differences between the baseline DFL and each gaze improvement method with regards to uncanniness, and thus \textbf{unable to reject the null hypothesis for H3.3}.

\section{Discussion}
\label{sec:discussion}

This work contributed to the growing discussion around human perception of face swaps.  We are the first to our knowledge to relate user perceptions of face swaps to uncanniness.  Our results verify that face swaps do generally elicit more uncanniness than real video counterparts.  Knowing this, there could be negative consequences in systems that implement face swaps for digital twins~\cite{caporusso_deepfakes_2021}, digital effects~\cite{naruniec_high-resolution_2020}, or privacy protection~\cite{lee_american_2021, zhu_deepfakes_2020, kuang_effective_2021, gafni_live_2019, sun_natural_2018, wilson_practical_2022}.  User experience in these systems could suffer if the generated media is perceived as uncanny or unsettling.

Based on our experiments, the proposed gaze improvement for face swapping using a pretrained gaze prediction model significantly decreases gaze error.  In a perceptual study, we have seen the proposed method impact users' perception of the eyes, specifically decreasing the prevalence of the eyes as a reliable factor when performing a deepfake detection task.

\revision[Here we have added a discussion point to highlight that while our original hypothesis was not verified, our results have merit and could be of use to the line of deepfake detection work using eye behaviors as a predictive feature.  This is based on reviewer 1's feedback, tying our original hypothesis back in to our findings.]{Based on our study verifying that generated face swaps can elicit uncanny feelings in viewers, and based on prior works analyzing user perceptions of face swaps~\cite{wohler_towards_2021, tahir_seeing_2021, gupta_eyes_2020}, we hypothesized that the eyes could attribute to perceived uncanniness, and that improving the quality of reconstructed gaze could alleviate uncanniness (H3.3).  While we did not produce sufficient evidence towards a decrease in uncanniness, our study has unveiled differences in human deepfake detection, particularly around viewer perception of the eyes.  With our method in place,}
\removal{While we do not have sufficient evidence of the proposed method decreasing uncanniness, we have found that with our method in place} the eyes are a less reliable feature for human deepfake detection.  These improvements could be quite beneficial to aid in the development of biometric deepfake classifiers that leverage gaze to label video as real or fake, by enabling the generation of higher quality training data~\cite{demir_where_2021, ciftci_fakecatcher_2020, ciftci_how_2020, li_ictu_2018}.

\subsection{Generalizing our Findings}

We have provided evidence to indicate that face swaps elicit uncanniness, but have not shown evidence of an uncanny valley.  The level of uncanniness experienced could depend on a number of factors, from source video quality or context to the algorithm used.  The analyzed dataset contains high quality data with high semantic pairing between subjects; in-the-wild video segments could have various pitfalls that both lower face swap quality and/ or increase the uncanniness experienced.  It would be a worthwhile future direction to more concretely map the uncanniness of face swaps given various factors.

DFL remains the most popular system online for visual effects face swapping and is continually being updated.  However, there are a growing number of face swapping publications specializing on arbitrary face swapping~\cite{nirkin_fsgan_2019, li_advancing_2020, chen_simswap_2020, liu_blendgan_2021} and super-resolution~\cite{liu_deepfacelab_2023, zhu_one_2021, wang_hififace_2021}.  These competing models are unlikely to have eliminated the presence of uncanniness, but should be investigated further to better quantify the effect across algorithms.  

\revision[Here we reiterate the novelty of our comparison with DFL+em, restate that the null hypothesis for comparison between the two is not rejected, and expand on the similarities between DFL+em and our DFL+Gaze.]{In addition to our proposed method (DFL+Gaze), which uses a pretrained gaze expert model to inform the training process, we also evaluate a pixel-matching method using masked eye regions (DFL+em), which, to our knowledge, is the only existing technique to prioritize eyes in face swapping.  We provide the first empirical evaluation of both approaches, seeing comparable effectiveness in gaze reconstruction accuracy.  However, we were unable to find statistically significant differences between our proposed method and the DFL+em baseline, and are unable to reject H2.2.  It is likely that the two approaches capture the same underlying information.  Our proposed gaze loss term incorporates gaze angle as a high-level feature.  This lessens the dependence on pixel-level matching of the eyes, possibly being more impactful at higher resolutions, but insignificant at the evaluated 128x128 resolution.}
\removal{Our proposed loss term achieves similar quantitative performance compared to DFL's native solution to the problem (which had not previously been quantified relative to the baseline).  These adjunct approaches likely capture much of the same information.}  However, it is important to note that the vast majority of face swapping approaches implement \textbf{neither} approach, so either could help to improve gaze representation.

\removal{Our proposed gaze loss term incorporates gaze angle as a high-level feature.  This lessens the dependence on pixel-level matching of the eyes, possibly being more impactful at higher resolutions.  The models in our evaluation generated faces at 128x128 pixels, yet} DFL supports up to 640x640 and other super-resolution systems can support up to 1024x1024~\cite{zhu_one_2021, naruniec_high-resolution_2020} given sufficient compute resources.  Our explicit focus on preserving gaze behavior could be applied to other face swapping pipelines.  Our implementation alters the optimization function but does not alter model architecture, meaning that our approach is fully additive.  Additionally, already-trained models could be further fine-tuned with this improvement in place.

While this analysis focused fully on gaze, a similar loss equation could be easily developed for other features, such as expression or head-pose matching.  Stacking multiple optimizations on the same network could improve overall quality.  Because our method's success is dependent on the pretrained gaze estimation model, the proposed approach will become more appealing as more advanced predictors are developed.  For example, current gaze predictors are prone to around 4\textdegree \space of prediction error, which likely bounds our method's performance.  When better performing predictors are created, our system will improve accordingly.  

\subsection{Limitations}

Our method uses the same pretrained network in training as our evaluation pipeline.  This opens up the possibility that our model could have learned to minimize the prediction error for L2CS-Net rather than generally improving gaze representations.  However, by observing how visually similar our results are to the native solution and the minimal differences in gaze errors, this concern is somewhat alleviated.  Our pipeline leveraged the pretrained gaze model to derive an angle error $\theta$.  If we had instead granted white-box access to the pretrained model's parameters, fitting to the gaze model would be more likely.

Our perceptual validation study provided some qualitative evidence towards the efficacy of our method.  However, the study exists on quite a small scale, only evaluating a small number of videos, all of which are specifically designed to enable high semantic quality in face swaps.  It would be worthwhile to further explore the potential improvements by focusing gaze during face swap training.  Results may be more significant on larger or more challenging stimuli sets, particularly in-the-wild videos which are subject to much lower data quality and more extreme head poses and gaze directions.
\section{Conclusion}
\label{sec:conclusion}

In this paper, we first identified the eyes as key features in previous studies revolving around the perception of face swaps.  Analyzing the training processes of face swapping models, we identified that typically training methodologies lack emphasis on the eye regions, which could be a cause for the perceptual differences seen.  

We then evaluate two low-cost training improvements that can aid models in better reconstructing eye gaze.  We evaluated a simple improvement using image masks to isolate the eyes and apply a more specific pixel-wise error metric.  We additionally proposed a novel loss equation which uses a pretrained gaze estimation model to guide training.  Both methods were successful in improving the accuracy of reconstructed gaze angles, yielding average improvements of 18.9\% and 21.2\%, respectively.

We chose to further evaluate our results qualitatively under the lens of the uncanny valley effect.  We are the first to our knowledge to analyze face swaps in terms of uncanniness.  We found evidence that face swaps do generally exist within the uncanny valley.  With our gaze improvements in place, we did not find significant improvements amongst uncanniness scores, but we did find that the eyes of our generated faces are less prominent as a decision factor for classifications as real or fake.

This advancement improves face swapping technology but is particularly promising for gaze-based deepfake detection; such an increase in fidelity will allow researchers to generate higher quality training datasets that will lead to better deepfake detection in real-world settings.  The additional findings regarding uncanniness will be an evaluation metric for future researchers implementing face swaps.

\section*{Acknowledgements}

All authors acknowledge funding from NIH R21 \textit{Protecting the Privacy of the Child Through Facial Identity Removal in Recorded Behavioral Observation Sessions} (Award R21MH123997).  Ethan Wilson acknowledges funding from the \textit{University of Florida Graduate School Preeminence Award} (GSPA) and \textit{Generation NEXT} scholarship (Award 1833908).

\bibliographystyle{template/cag-num-names}
\bibliography{misc_biblio,zotero_biblio}

\end{document}